\documentclass[letterpaper,10pt,conference]{ieeeconf}

\IEEEoverridecommandlockouts
\usepackage{graphicx}
\usepackage{amsmath}
\usepackage{amssymb}
\usepackage{booktabs}
\usepackage{multirow}
\usepackage{xcolor}
\usepackage{cite}
\usepackage{url}
\usepackage{placeins}
\usepackage{float}

\makeatletter
\newcommand{\fixedfigurecaption}[1]{%
  \refstepcounter{figure}%
  \@makecaption{\fnum@figure}{#1}%
}
\makeatother

\makeatletter
\def\section{\@startsection{section}{1}{\z@}%
  {0.9ex plus 0.4ex minus 0.2ex}%
  {0.3ex plus 0.2ex minus 0ex}%
  {\normalfont\normalsize\centering\scshape}}%
\def\subsection{\@startsection{subsection}{2}{\z@}%
  {0.9ex plus 0.4ex minus 0.2ex}%
  {0.3ex plus 0.2ex minus 0ex}%
  {\normalfont\normalsize\itshape}}%
\makeatother

\title{\LARGE \bf
PanoGraphGS: Unifying Panoptic Segmentation and Hierarchical 3D Scene Graphs with Gaussian Splatting
}

\author{
Jinghan Gao$^{*}$, Siyu Wang$^{*}$, Dongyu Feng, Yiyuan Pan, Rong Xiong%
\thanks{$^{*}$Jinghan Gao and Siyu Wang contributed equally to this work.}%
\thanks{All authors are with the Zhejiang Humanoid Robot Innovation Center, Ningbo, China.}%
\thanks{Corresponding author: Yiyuan Pan (email: panyiyuan@zju.edu.cn.)}%
}

\begin{document}

\raggedbottom

\maketitle
\thispagestyle{empty}
\pagestyle{empty}


\begin{abstract}
3D scene graphs support language-guided navigation by organizing physical
entities and their containment and spatial relations into hierarchical maps.
Building such structured maps requires a scene representation that consistently
distinguishes semantic regions and object instances across viewpoints. To this
end, we present PanoGraphGS, a unified framework that integrates panoptic 3D
Gaussian representation learning with hierarchical 3D scene graph construction.
The framework learns semantic and instance features on shared Gaussian
geometry. Prototype-based semantic learning and association-based instance
learning produce consistent panoptic predictions across viewpoints. The learned
representation supports novel-view panoptic segmentation and direct
Gaussian-level panoptic decoding. A semantic-aware scene graph module further
recovers a floor-room-stuff-object hierarchy with containment and spatial
relations.  We further release NaviGS, a real-world dataset for panoramic scene
understanding. The dataset provides 3DGS, mesh, RGB, semantic, and depth data. For 3D scene graph identification, PanoGraphGS
outperforms the strongest evaluated baseline in object-node accuracy
by more than 50\% points on average across Replica and
NaviGS. 
PanoGraphGS also achieves
state-of-the-art performance among the evaluated methods in 2D
panoptic segmentation. These results demonstrate the effectiveness of PanoGraphGS
in both hierarchical 3D scene
understanding and 2D panoptic segmentation. The project website is available at:
\url{https://github.com/gaojh0333/PanoGraphGS}.
\end{abstract}

\section{INTRODUCTION}

Panoptic 3D scene reconstruction is fundamental to language-guided robotic
navigation. Successful navigation requires embodied agents to recognize
semantic categories, maintain persistent instance identities, and reason about
spatial organization in a globally consistent map. Such
tasks require reasoning over nested containment and spatial relations. A 3D
scene graph formalizes this organization by representing physical entities as
nodes and their containment and spatial relations as
edges~\cite{armeni2019scenegraph,wald2020learning3dssg}.
3D Dynamic Scene
Graphs~\cite{rosinol20203ddsg}, Hydra~\cite{hughes2022hydra}, and
HOV-SG~\cite{werby2024hovsg}, have advanced structured spatial mapping.
However, these pipelines commonly rely on per-pixel RGB-D observations or
depth-derived geometry. Incomplete or noisy depth can degrade object-extent
estimation, cross-view association, room partitioning, and spatial-relation
inference. Moreover, existing methods typically construct scene graphs and
dense panoptic representations separately, requiring an additional stage to
associate panoptic predictions with graph entities.

\begin{figure}[!t]
    \centering
    \includegraphics[width=0.9\linewidth]{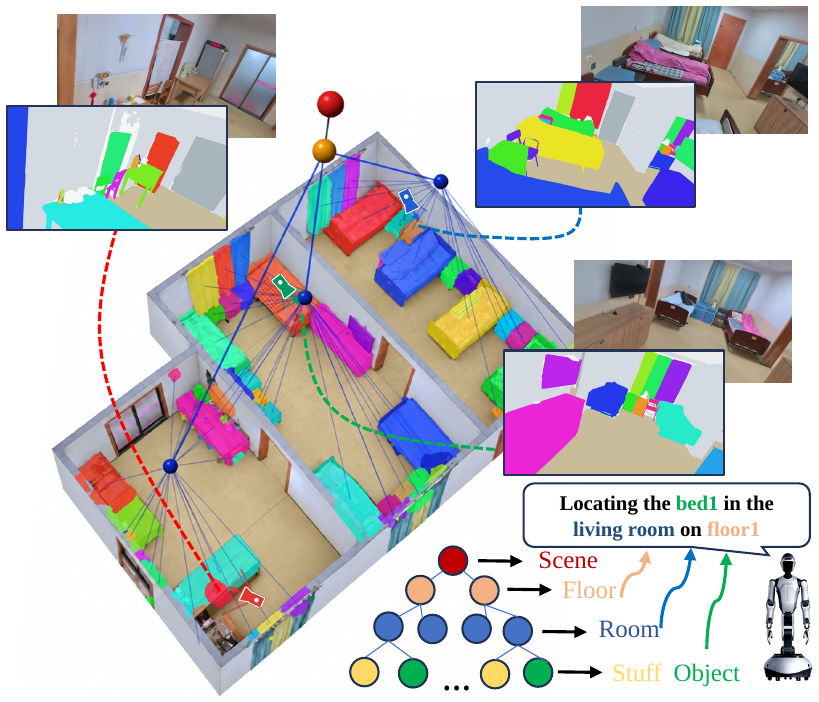}
    \caption{\textbf{Results of PanoGraphGS.} PanoGraphGS is the first method to construct \textit{floor--room--stuff--object} scene graphs for language-guided robotic navigation, improving object accuracy while reducing stuff interference. It also enables cross-frame consistent panoptic segmentation with persistent instance identities.}
    \label{fig:resthome15}
\end{figure}

\begin{figure*}[!t]
    \centering
    \includegraphics[width=0.99\textwidth]{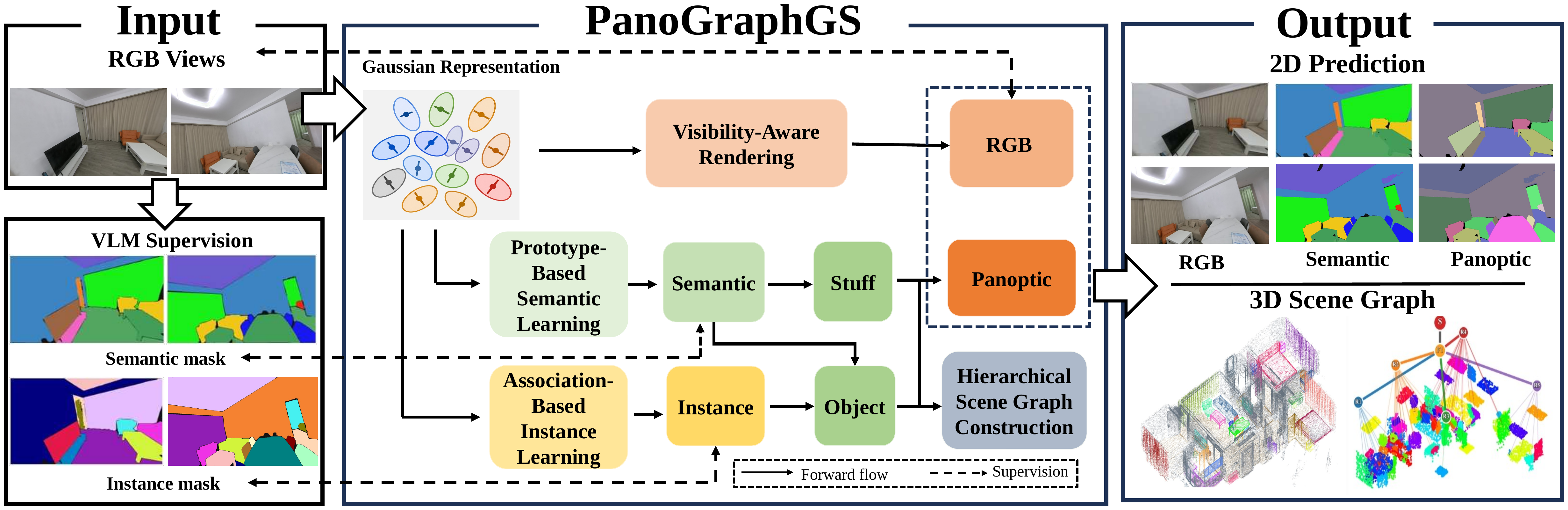}
    \caption{\textbf{Overview of PanoGraphGS.} From multi-view RGB inputs and SAM3-generated masks~\cite{carion2025sam3segmentconcepts} to a shared panoptic Gaussian representation, panoptic predictions, and hierarchical 3D scene graph construction. The center illustrates joint semantic and instance learning on shared Gaussian geometry.}
    \label{fig:framework_overview}
\end{figure*}

These limitations motivate a unified 3D representation that preserves
geometry, semantic categories, and persistent instance identities while being
learned directly from multi-view RGB observations. 3D Gaussian Splatting
(3DGS)~\cite{kerbl20233dgs} provides such a representation through explicit,
spatially anchored Gaussian primitives and differentiable rendering.
Visibility-aware rendering lifts multi-view semantic and instance supervision
into a shared 3D field. The learned features can be rendered for novel-view
panoptic prediction or decoded directly from Gaussians for 3D entity
abstraction. Unified-Lift~\cite{zhu2025unifiedlift} learns Gaussian- and
object-level representations from multi-view segmentation cues. More recently,
GaussianGraph~\cite{wang2025gaussiangraph} extends 3DGS toward scene graph
generation by clustering Gaussian primitives and associating them with CLIP
features and spatial relations extracted from 2D foundation models. However,
its reasoning mainly focuses on object-level spatial relations and relies on
high-dimensional vision--language features for semantic association, while
clustering may discard fine-grained spatial information. Explicit
identification, matching, and hierarchical association across the
floor-room-stuff-object hierarchy remain unaddressed. Consequently, dense
panoptic prediction and hierarchical scene organization remain loosely
integrated. Establishing globally consistent panoptic predictions within a
Gaussian representation therefore presents two challenges. First,
viewpoint-dependent visibility and occlusion produce nonuniform multi-view
supervision. Second, the same physical object may receive different view-local
pseudo-instance identifiers across views, causing conflicting identity
supervision without cross-view association. Gaussian-level labels also do not
explicitly encode the floor-room-stuff-object hierarchy or its containment and
spatial relations. Recovering this structure requires joint reasoning over
Gaussian semantics, persistent instance identities, and reconstructed geometry.

To address these challenges, we propose PanoGraphGS, an end-to-end framework
that integrates panoptic 3D Gaussian representation learning with hierarchical
3D scene graph construction. PanoGraphGS jointly learns semantic and instance
features anchored to shared Gaussian geometry. The semantic branch combines
residual feature projection and category prototypes with confidence-weighted
supervision, while the instance branch associates view-local pseudo-labels with
persistent Gaussian instance identities to ensure consistent object
correspondence across views. As illustrated in Fig.~\ref{fig:resthome15}, the
resulting panoptic Gaussian representation supports both panoptic segmentation
and direct Gaussian-level decoding. The
semantic-aware hierarchical reasoning module combines structural semantics with
Gaussian geometry to recover the \textit{floor-room-stuff-object} hierarchy
and its containment and spatial relations, enabling structured spatial queries
such as \textit{locating the bed1 in the living room on floor1}. CLIP~\cite{radford2021clip} is further used to assign functional
categories to rooms. Our main contributions are:

\begin{itemize}
\item We propose Prototype-Based Semantic Learning, which jointly optimizes 2D rendered and 3D Gaussian features in a unified semantic space, overcoming the limitation of optimizing features in only the 2D or 3D domain.

\item We propose Association-Based Instance Learning, which associates view-local instance masks with persistent identities for cross-frame consistent panoptic prediction, with confidence-aware supervision reducing the impact of unreliable labels.

\item We propose a semantic-aware hierarchical 3D scene graph construction method that organizes learned panoptic Gaussians into a \textit{floor-room-stuff-object} hierarchy without additional training or RGB-D inputs.

\item We achieve state-of-the-art performance in both 2D panoptic segmentation and 3D scene graph construction.
\end{itemize}

\section{RELATED WORK}
\label{sec:related_work}



\subsection{3D Gaussian Segmentation}

3DGS-based segmentation has progressed from open-vocabulary semantics and instance grouping to panoptic understanding. Open-vocabulary methods embed vision--language features into 3D Gaussian primitives. LangSplat~\cite{qin2024langsplat} learns language-aligned Gaussian features, while OpenGaussian~\cite{wu2024opengaussian} and OVGaussian~\cite{chen2025ovgaussian} extend this paradigm to point-level open-vocabulary segmentation. Mask-lifting methods transfer 2D masks and instance information into Gaussian representations. Gaussian Grouping~\cite{ye2024gaussiangrouping} learns identity-aware Gaussian features for object grouping, while Unified-Lift~\cite{zhu2025unifiedlift} combines Gaussian- and object-level representations with association learning for 3D instance segmentation. More recently, PLGS~\cite{wang2025plgs} jointly learns semantic and instance attributes from multi-view supervision, and PanoSLAM~\cite{chen2025panoslam} integrates panoptic Gaussian learning with sequential RGB-D mapping. Similarly, PanoGS~\cite{zhai2025panogs} combines open-vocabulary semantics with instance segmentation, while PanopticSplatting~\cite{xie2025panopticsplatting} introduces query-guided cross-view label refinement for panoptic segmentation. Despite these advances, existing methods mainly focus on novel-view panoptic prediction, with feature optimization restricted to either the 2D or 3D domain. In contrast, PanoGraphGS jointly optimizes 2D and 3D features, enabling both panoptic segmentation and direct Gaussian-level decoding.

\begin{figure*}[!t]
    \centering
    \includegraphics[width=0.95\textwidth]{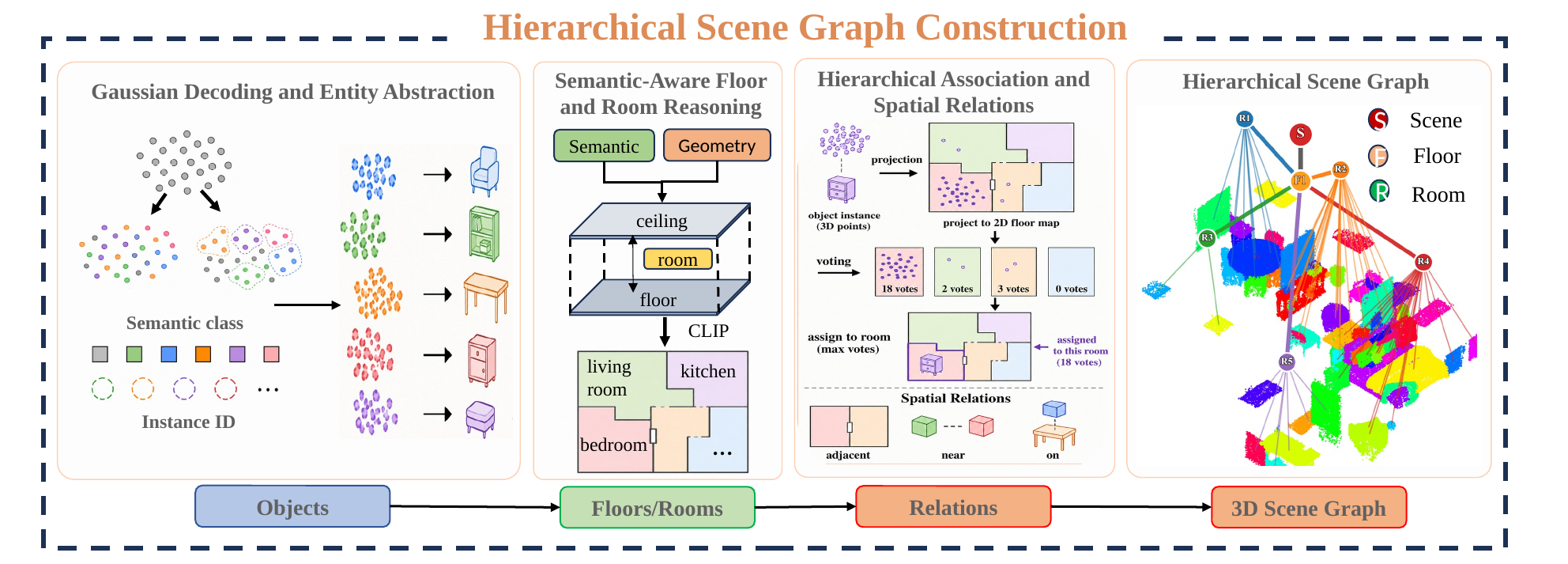}
    \setlength{\abovecaptionskip}{1pt}
    \caption{Hierarchical 3D scene graph construction from panoptic Gaussians. 
    }
    \label{fig:scenegraph_construction}
\end{figure*}

\subsection{Hierarchical 3D Scene Graphs}
Armeni \emph{et al.}~\cite{armeni2019scenegraph} first organize buildings, rooms, and objects into a multi-layer graph, while 3DSSG~\cite{wald2020learning3dssg} learns object relations from indoor reconstructions. 3DDGS~\cite{rosinol20203ddsg} integrate geometry with objects, places, and agents, Hydra~\cite{hughes2022hydra} enables online hierarchical mapping, and HOV-SG~\cite{werby2024hovsg} introduces an open-vocabulary floor-room-object hierarchy. GaussianGraph~\cite{wang2025gaussiangraph} novelly constructs structured scene graphs directly from Gaussians. However, these approaches commonly rely on per-pixel RGB-D observations for geometry and primarily focus on object-level recognition. In contrast, PanoGraphGS leverages learned semantic and instance features together with Gaussian geometry to recover a hierarchical floor-room-stuff-object scene graph with explicit containment and spatial relations without training.

\section{METHOD}

\subsection{Overall}

As illustrated in
Figure~\ref{fig:framework_overview}, we employ SAM3-generated~\cite{carion2025sam3segmentconcepts} masks from posed RGB images as supervision, which may contain noisy
labels and inconsistent instance IDs across frames. We first augment the
Gaussians with semantic and instance features (Section~\ref{subsec:3dgs_enhancement}).
We jointly optimize the appearance, semantic, and instance branches. The prototype-based semantic learning predicts semantic categories (Section~\ref{subsec:semantic_learning}),
while the association-based instance learning aligns view-local
instance masks with persistent instance identities
(Section~\ref{subsec:instance_learning}). The joint supervision strategy is
described (Section~\ref{subsec:supervision}). During panoptic decoding, stuff
predictions are obtained from the semantic branch, whereas object predictions
combine semantic and instance. The resulting
panoptic Gaussian representation supports novel-view 2D panoptic segmentation
and direct Gaussian-level panoptic decoding. Finally, the semantic-aware
hierarchical scene graph construction module organizes the learned panoptic
Gaussians into a \textit{floor--room--stuff--object} hierarchy with containment
and spatial relations (Section~\ref{subsec:scene_graph_construction}).

\subsection{3DGS Enhancement}
\label{subsec:3dgs_enhancement}

Following 3DGS~\cite{kerbl20233dgs}, the scene is represented by anisotropic
Gaussians $\mathcal G=\{g_i\}_{i=1}^{N}$.
Each Gaussian $g_i$ is parameterized by a 3D mean $\boldsymbol\mu_i$, scale
$\mathbf s_i$, rotation $\mathbf R_i$, opacity $\alpha_i$, and view-dependent
color coefficients $\mathbf c_i$, together with independent semantic and
instance features
$\mathbf F_i^{\mathrm{sem}},\mathbf F_i^{\mathrm{ins}}\in\mathbb R^D$
($D=64$). The two feature fields share Gaussian support and visibility but
use separate prediction heads. Here, $i$ indexes Gaussians, $p$ indexes
pixels, $b\in\{\mathrm{sem},\mathrm{ins}\}$ denotes the feature branch, and
$c$, $r$, and $k$ denote a semantic category, a view-local pseudo-instance
identifier, and a persistent global instance identity, respectively.
RGB and branch-specific features are rendered with the same visibility-aware
alpha compositing. For Gaussians ordered from front to back at pixel $p$, the
compositing weights and rendered outputs are:
\begin{equation}
 w_{i,p}=\alpha_{i,p}\!\prod_{j<i}(1-\alpha_{j,p}),
\end{equation}
\begin{equation}
 \hat{\mathbf C}_p=\sum_i w_{i,p}\mathbf c_{i,p},  
 \mathbf f_p^b=\sum_i w_{i,p}\mathbf F_i^b,  
 b\in\{\mathrm{sem},\mathrm{ins}\},
 \label{eq:feature_rendering}
\end{equation}
where $\alpha_{i,p}=\alpha_i\mathcal K_i(p)$ is the opacity,
$\mathcal K_i(p)$ is the projected Gaussian kernel at $p$, and
$\mathbf c_{i,p}$ is the view-dependent color of $g_i$. The outputs
$\hat{\mathbf C}_p$ and $\mathbf f_p^b$ denote the rendered RGB color and
feature of branch $b$, respectively. This shared rendering lifts
multi-view 2D supervision to Gaussian-level features.

\begin{figure*}[!t]
    \centering
    \includegraphics[width=0.9\textwidth]
    {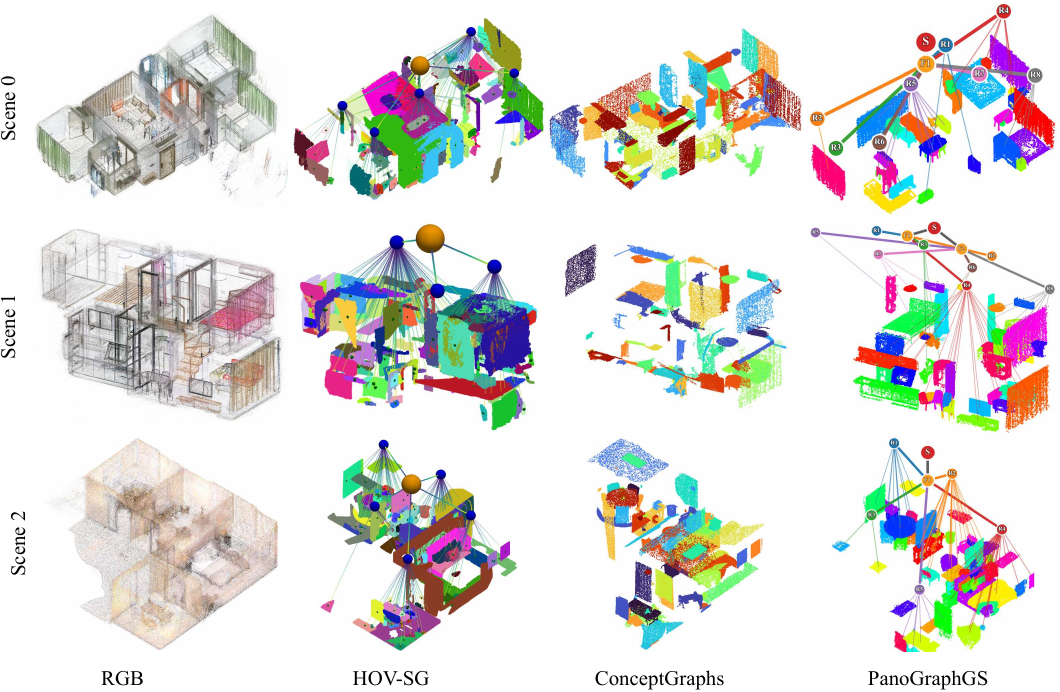}
    \caption{3D scene graph visualization on NaviGS and Replica.}
    \label{fig:scenegraph_visualization}
\end{figure*}

\subsection{Prototype-Based Semantic Learning}
\label{subsec:semantic_learning}

The semantic branch maintains a learnable prototype
$\mathbf q_c\in\mathbb R^D$ for each semantic category $c$. Each
$\mathbf q_c$ is randomly initialized. We introduce a residual-projection MLP
shared between rendered 2D pixel features and 3D Gaussian-level
features. This MLP unifies image-space semantic supervision with
Gaussian-level semantic decoding and lifts 2D panoptic information into 3D.
The discrepancy between the two spaces can produce noisy semantic
predictions when 2D supervision is directly transferred to 3D Gaussian
representations. This design mitigates the discrepancy. Specifically,
the residual projection maps the rendered semantic feature
$\mathbf f_p^{\mathrm{sem}}$ to an embedding. Its similarity to each
prototype determines the semantic logit:
\begin{equation}
\begin{aligned}
 \tilde{\mathbf f}_p^{\mathrm{sem}}
 &= \mathbf f_p^{\mathrm{sem}}
 + \operatorname{MLP}_{\mathrm{sem}}(\mathbf f_p^{\mathrm{sem}}),\\
 \mathbf z_p^{\mathrm{sem}}
 &= \mathcal N\!\left(
 \operatorname{LN}\!\left(\tilde{\mathbf f}_p^{\mathrm{sem}}\right)\right),\\
 \ell_{p,c}^{\mathrm{sem}}
 &= (\mathbf z_p^{\mathrm{sem}})^\top \mathcal N(\mathbf q_c),
\end{aligned}
\label{eq:semantic_head}
\end{equation}
where $\operatorname{MLP}_{\mathrm{sem}}$ is a two-layer semantic projection,
$\operatorname{LN}$ is layer normalization, and
$\mathcal N(\mathbf x)$ is
$\ell_2$ normalization, $\tilde{\mathbf f}_p^{\mathrm{sem}}$ is the
residual-projected feature, $\mathbf z_p^{\mathrm{sem}}$ is its normalized
embedding, and $\ell_{p,c}^{\mathrm{sem}}$ is the logit for category $c$.
For Gaussian-level semantic decoding,
$\mathbf f_p^{\mathrm{sem}}$ is replaced with $\mathbf F_i^{\mathrm{sem}}$ in
Eq.~\eqref{eq:semantic_head}. This allows rendered and Gaussian-level features
to share the same classifier. Both the semantic embedding and class prototypes
are $\ell_2$-normalized.

\subsection{Association-Based Instance Learning}
\label{subsec:instance_learning}


SAM3-generated instance labels are inconsistent across views. The same physical
object may receive different identifiers and is thus treated as a pseudo-label.
To address this ambiguity, the instance branch maintains $K$ learnable global
instance prototypes ${\mathbf u_k}_{k=1}^{K}$ and associates view-local
instance regions with persistent global identities. Given the rendered
instance feature $\mathbf f_p^{\mathrm{ins}}$, the global instance logits and
probabilities are defined as:
\begin{equation}
\ell_{p,k}^{\mathrm{ins}}
= \mathbf u_k^\top \mathbf f_p^{\mathrm{ins}},
\qquad
P_{p,k}^{\mathrm{ins}}
= \operatorname{softmax}(\boldsymbol\ell_p^{\mathrm{ins}})_k,
\label{eq:instance_logits}
\end{equation}
For a view-local pseudo-instance $r$, pixelwise probabilities
are aggregated over its region to define the association cost:
\begin{equation}
 A_{r,k} = -\sum_{p:y_p^{\mathrm{ins}}=r} P_{p,k}^{\mathrm{ins}},
\end{equation}
and the one-to-one assignment between local pseudo-instances and
global instance identities is determined by
\begin{equation}
 \pi_v^*=
 \arg\min_{\pi_v}
 \sum_{r\in\mathcal R_v} A_{r,\pi_v(r)},
 \label{eq:instance_assignment}
\end{equation}
 where $\mathbf u_k$ is the learnable prototype of global instance identity $k$,
$y_p^{\mathrm{ins}}$ is the view-local pseudo-instance supervision at pixel
$p$, $\mathcal R_v$ is the set of pseudo-instance identifiers in view $v$,
and $\pi_v^*$ is the optimal one-to-one local-to-global association.
The Hungarian algorithm~\cite{kuhn1955hungarian} solves this assignment.
The resulting association aligns view-local pseudo-labels with persistent
Gaussian instance identities, rather than directly matching pairs of camera
views. Gaussian-level instance identities are decoded with the same prototype
scores by replacing $\mathbf f_p^{\mathrm{ins}}$ with
$\mathbf F_i^{\mathrm{ins}}$ in Eq.~\eqref{eq:instance_logits}.

\subsection{End-to-end Learning}
\label{subsec:supervision}
To jointly optimize the appearance, semantic, and instance branches, we
introduce a unified objective combining appearance, semantic, and instance
supervision. It is formulated as:
\begin{equation}
\mathcal L_{\rm joint} =
\mathcal L^{\rm rgb}
+
\mathcal L^{\rm sem}
+
\mathcal L^{\rm ins},
\label{eq:joint_loss}
\end{equation}
where $\mathcal L^{\rm rgb}$, $\mathcal L^{\rm sem}$, and $\mathcal L^{\rm ins}$ denote the appearance, semantic, and instance losses, respectively, constraining appearance consistency and semantic and instance supervision on the shared representation.

\textbf{Appearance loss.} Following standard 3DGS, the
appearance loss combines $L1$ loss and $L_{SSIM}$ to balance
pixel-wise appearance reconstruction and structural similarity:
\begin{equation}
\mathcal L^{\rm rgb} = (1-\mu )L1 + \mu L_{SSIM}
\end{equation}
where $\mu$ = 0.2.

\begin{table*}[!t]
\centering
\caption{Hierarchical scene graph comparison on Replica and NaviGS datasets. }
\label{tab:scenegraph_three_metrics}
\setlength{\tabcolsep}{4pt}
\renewcommand{\arraystretch}{1}
\begin{tabular}{l|ccc|ccc}
\toprule
\multirow{2}{*}{Method}
& \multicolumn{3}{c|}{Replica}
& \multicolumn{3}{c}{NaviGS} \\
& Floor Acc. (\%) & Room Acc. (\%) & Object Acc. (\%)
& Floor Acc. (\%) & Room Acc. (\%) & Object Acc. (\%) \\
\midrule
KeySG~\cite{werby2025keysg}
& 100.00 & 22.50 & 17.27 & 90.00 & 35.00 & 5.25 \\
HOV-SG~\cite{werby2024hovsg}
& 100.00 & 87.50 & 21.72 & 90.00 & 66.67 & 14.57 \\
OP3DSG~\cite{kim2026op3dsg}
& -- & -- & 35.91 & -- & -- & 16.97 \\
BBQ~\cite{linok2025bbq}
& -- & -- & 33.08 & -- & -- & 26.25 \\
ConceptGraphs~\cite{gu2023conceptgraphs}
& -- & -- & 39.32 & -- & -- & 30.50 \\
\midrule
PanoGraphGS (Ours)
& \textbf{100.00} & \textbf{100.00} & \textbf{89.90}
& \textbf{100.00} & \textbf{100.00} & \textbf{98.36} \\
\bottomrule
\end{tabular}
\end{table*}

\textbf{Confidence-weighted mechanism.} To mitigate unreliable semantic and
instance supervision near object boundaries and occlusions, we introduce a
confidence-aware weighting mechanism for pixel-wise supervision. For each task
$t\in{\mathrm{sem},\mathrm{ins}}$, let $\mathcal{P}_k^t$ denote pixels assigned
to category or instance $k$ in a training view. We compute a task-specific
feature center $\mathbf{m}_k^t$ and a dispersion-dependent temperature
$\tau_k^t$. The confidence of pixel $p$ for task $t$ is defined as
\begin{equation}
\begin{aligned}
 \mathbf m_c^{\mathrm{sem}} &= \frac{1}{|\mathcal P_c|}\sum_{p\in\mathcal P_c}
 \mathcal N(\mathbf f_p^{\mathrm{sem}}),\\
 \gamma_p^{\mathrm{sem}} &= \frac{\exp(\mathcal N(\mathbf f_p^{\mathrm{sem}})^\top\mathbf m_{y_p^{\mathrm{sem}}}^{\mathrm{sem}}/\tau_{y_p^{\mathrm{sem}}})}
 {\sum_{c\in\mathcal C_v}\exp(\mathcal N(\mathbf f_p^{\mathrm{sem}})^\top\mathbf m_c^{\mathrm{sem}}/\tau_c)+\epsilon},
\end{aligned}
\label{eq:confidence}
\end{equation}
where $y_p^t$ denotes the supervision label at pixel $p$, and
$\mathcal C_v^t$ denotes the set of semantic categories or instance identities
present in view $v$. The temperature $\tau_k^t$ adjusts confidence according
to the within-class or within-instance feature dispersion. The confidence
$\gamma_p^t$ measures supervision reliability relative to competing categories
or instance identities in the current view. It suppresses unreliable samples
in both the semantic and instance objectives. The task- and
class/instance-dependent factor $\tau_k^t$ is used only for confidence
weighting. It is not involved in the semantic or instance classifiers.

\textbf{Semantic loss.}
The semantic objective is
\begin{equation}
\mathcal L^{\mathrm{sem}}=\mathcal L_{\rm cls}^{\mathrm{sem}}+
\lambda_{\rm con}\mathcal L_{\rm con}^{\mathrm{sem}}+
\lambda_{\rm clu}\mathcal L_{\rm clu}^{\mathrm{sem}}+
\lambda_{\rm reg}\mathcal L_{\rm reg}^{\mathrm{sem}}.
\label{eq:semantic_loss}
\end{equation}
where $\lambda_{\rm con}=\lambda_{\rm reg}=10^{-6}$ and
$\lambda_{\rm clu}=0.1$. The terms $\mathcal L_{\rm cls}^{\mathrm{sem}}$,
$\mathcal L_{\rm con}^{\mathrm{sem}}$,
$\mathcal L_{\rm clu}^{\mathrm{sem}}$, and
$\mathcal L_{\rm reg}^{\mathrm{sem}}$ denote confidence-weighted classification, region-level contrastive learning,
prototype alignment, and feature-norm regularization, respectively. The
classification term uses label-smoothed cross-entropy~\cite{szegedy2016inception}.
Confidence-weighted defined in Eq.~\eqref{eq:confidence} reduces the
influence of unreliable pseudo-labels and improves robustness to noisy
supervision~\cite{reed2015noisy}. Region-level contrastive term follows
supervised contrastive learning~\cite{khosla2020supervised} and encourages
coherent features within the same semantic regions. Prototype alignment
promotes consistency between reliable semantic embeddings and category
prototypes~\cite{snell2017prototypical}. Feature-norm regularization stabilizes shared embedding space.

\textbf{Instance loss.}
Analogous to the semantic branch, the instance objective uses confidence-aware
supervision and region-level contrastive learning on rendered instance features.
View-local instance regions provide supervision, while the association in
Eq.~\eqref{eq:instance_assignment} maps them to persistent global identities
for instance classification and prototype alignment. The resulting instance
objective $\mathcal L^{\mathrm{ins}}$ follows the same four-term formulation
as the semantic objective.

\subsection{Hierarchical Scene Graph Construction}
\label{subsec:scene_graph_construction}

As showned in Fig.~\ref{fig:scenegraph_construction}, the panoptic Gaussian representation is organized into a hierarchical 3D scene
graph following a floor-room-stuff-object hierarchy. Floor and room nodes
capture spatial structure, while stuff and object nodes represent structural
semantics and persistent instance identities, respectively. Node attributes
are grounded in the learned panoptic features and Gaussian geometry, with
directed edges encoding hierarchical containment and spatial relations.

\textbf{Gaussian Decoding and Entity Abstraction.}
The Gaussian representation is first decoded into semantic and instance-aware
entities.
\begin{equation}
\mathcal{G}=\left \{ {(\mathbf{x}_i,\hat{s}_i,\hat{u}_i)} \right \} _{i=1}^{N},
\end{equation}
where $\mathbf{x}_i$ denotes the 3D position of the $i$-th Gaussian, and
$\hat{s}_i$ and $\hat{u}_i$ denote its predicted semantic category and
persistent instance identity, respectively. For an object with instance
identity $k$, its supporting Gaussian set is defined as
\begin{equation}
\mathcal{G}_k^{object}=\left \{ {i \mid \hat{u}_i)=k} \right \} ,
\end{equation}
whereas structural stuff entities are obtained by grouping Gaussians according
to their semantic categories:
\begin{equation}
\mathcal{G}_c^{stuff}=\left \{ {i \mid \hat{s}_i)=c} \right \} ,
\end{equation}
Thus, the dense Gaussian representation is abstracted into a compact set of
entities $\mathcal{V}$:
\begin{equation}
\mathcal{V}_{\mathrm{pan}} = 
\mathcal{G}_c^{stuff}
\cup
\mathcal{V}_{\mathrm{object}},
\end{equation}
where each entity retains its semantic identity and geometric support from the
underlying Gaussians.

\textbf{Semantic-Aware Floor and Room Reasoning.}
The floor-room hierarchy is inferred by jointly exploiting semantic
predictions and the spatial distribution of the corresponding Gaussians.
Specifically, floor and ceiling semantics provide the primary evidence for
identifying floor regions, with the space between the corresponding floor and
ceiling Gaussians defining a floor and their vertical separation determining
its height. The vertical distribution of these supporting Gaussians further
provides complementary geometric cues. The floor reasoning process is
formulated as
\begin{equation}
\left\{
\mathcal{G}_{\mathrm{floor}}^{stuff},
\mathcal{G}_{\mathrm{ceiling}}^{stuff}
\right\}
\xrightarrow{}
\mathcal{V}_{\mathrm{floor}},
\end{equation}
Within each identified floor, structural semantic categories and their spatial
organization are used to delineate room-level regions, with walls, doors, and
windows providing explicit semantic boundary cues for room partitioning:
\begin{equation}
\mathcal{V}_{\mathrm{floor}}
\xrightarrow{
\mathcal{G}_{\mathrm{wall}}^{stuff},
\mathcal{G}_{\mathrm{door}}^{stuff},
\mathcal{G}_{\mathrm{window}}^{stuff},
\cdots
}
\mathcal{V}_{\mathrm{room}},
\end{equation}
For each room, RGB observations are rendered from camera poses within the
corresponding room and used with CLIP~\cite{radford2021clip} to assign a
functional semantic label:
\begin{equation}
\hat{y}^{\mathrm{room}}_r =
\operatorname*{arg\,max}_{c \in \mathcal{C}_{\mathrm{room}}} S(r,c),
\label{eq:room_function_label}
\end{equation}
where $S(r,c)$ denotes the aggregated semantic similarity between the
observations associated with room $r$ and candidate functional category $c$, which provides functional information while preserving the
geometry of the room.

\begin{table*}[!t]
\centering
\caption{Comparison of PanoGraphGS with state-of-the-art algorithms on Scannet++, Replica and NaviGS benchmarks Panoptic segmentation.
The best and second-best results are indicated in bold and underlined, respectively.}
\label{tab:main_quantitative}
\setlength{\tabcolsep}{15pt}
\renewcommand{\arraystretch}{1}
\begin{tabular}{l|cc|cc|cc}
\toprule
\multirow{2}{*}{Model}
& \multicolumn{2}{c|}{ScanNet++}
& \multicolumn{2}{c|}{Replica}
& \multicolumn{2}{c}{NaviGS} \\
& mIoU & mBIoU & mIoU & mBIoU & mIoU & mBIoU \\
\midrule
Gaussian Grouping~\cite{ye2024gaussiangrouping}
& 15.67 & 13.03 & 23.94 & 16.98 & 56.91 & 35.86 \\
ObjectGS~\cite{zhu2025objectgs}
& 8.91 & 6.41 & 12.22 & 5.39 & 43.77 & 26.69 \\
Trace3D~\cite{shen2025trace3d}
& 30.22 & 27.54 & 54.90 & 48.50 & 67.51 & 44.75 \\
Unified-Lift~\cite{zhu2025unifiedlift}
& \underline{38.10} & \underline{35.92}
& \underline{63.91} & \underline{59.69}
& \underline{78.48} & \underline{57.84} \\
PanopticRecon++~\cite{yu2025panopticreconpp}
& 31.93 & 29.31 & 54.32 & 43.51 & 49.36 & 25.58 \\
\midrule
PanoGraphGS (Ours)
& \textbf{41.06} & \textbf{37.89}
& \textbf{76.73} & \textbf{70.51}
& \textbf{83.19} & \textbf{64.39} \\
\bottomrule
\end{tabular}
\end{table*}

\textbf{Hierarchical Association and Spatial Relations.}
We first associate the extracted objects with their corresponding rooms
through voting. For each object $e$, we count its
supporting 3D points within the bounding box of each candidate room $r$. The
room receiving the most votes is selected:
\begin{equation}
r(e)= \arg\max_{r \in \mathcal{V}_{\mathrm{room}}}
\left| \left\{ \mathbf{x}_i \in \mathcal{G}_e \;\middle|\; \mathbf{x}_i \in \mathrm{BBox}(r) \right\} \right|,
\label{eq:room_assignment}
\end{equation}
where $\mathcal{G}_e$ denotes the set of 3D Gaussians supporting object $e$,
and $\mathrm{BBox}(r)$ represents the bounding box of room $r$. This
point-level voting strategy assigns each object to the room containing the
largest portion of its 3D support.

The extracted entities are finally organized into a hierarchical scene graph
$\mathcal{S}=(\mathcal{V},\mathcal{E})$, where the edge set $\mathcal{E}$
comprises hierarchical containment and spatial relations:
\begin{equation}
    \mathcal{E} = \mathcal{E}_{\mathrm{contain}} \cup \mathcal{E}_{\mathrm{spatial}},
    \label{eq:scene_graph_edges}
\end{equation}
Specifically, the containment relations establish the hierarchical structure
of the scene:
\begin{equation}
    \mathcal{E}_{\mathrm{contain}} = 
    \left\{ (f,r) \mid r \subset f \right\} \cup
    \left\{ (r,e) \mid r=r(e),\ e \in \mathcal{V}_{\mathrm{object}} \right\}.
    \label{eq:containment_relations}
\end{equation}
where $r \subset f$ denotes that room $r$ is spatially contained within floor
$f$. The spatial relations $\mathcal{E}_{\mathrm{spatial}}$ further describe
the topological and geometric organization among entities within the same
hierarchical level. Consequently, the final graph preserves both the metric
grounding of the Gaussian representation and its high-level semantic structure,
without introducing an additional learned graph predictor.

\section{EXPERIMENTS}

We first compare PanoGraphGS on 3D scene graph construction and 2D panoramic segmentation, with qualitative visualizations provided for both, followed by ablation studies of its key components.


\begin{figure*}[!t]
    \centering
    \includegraphics[width=0.95\textwidth]{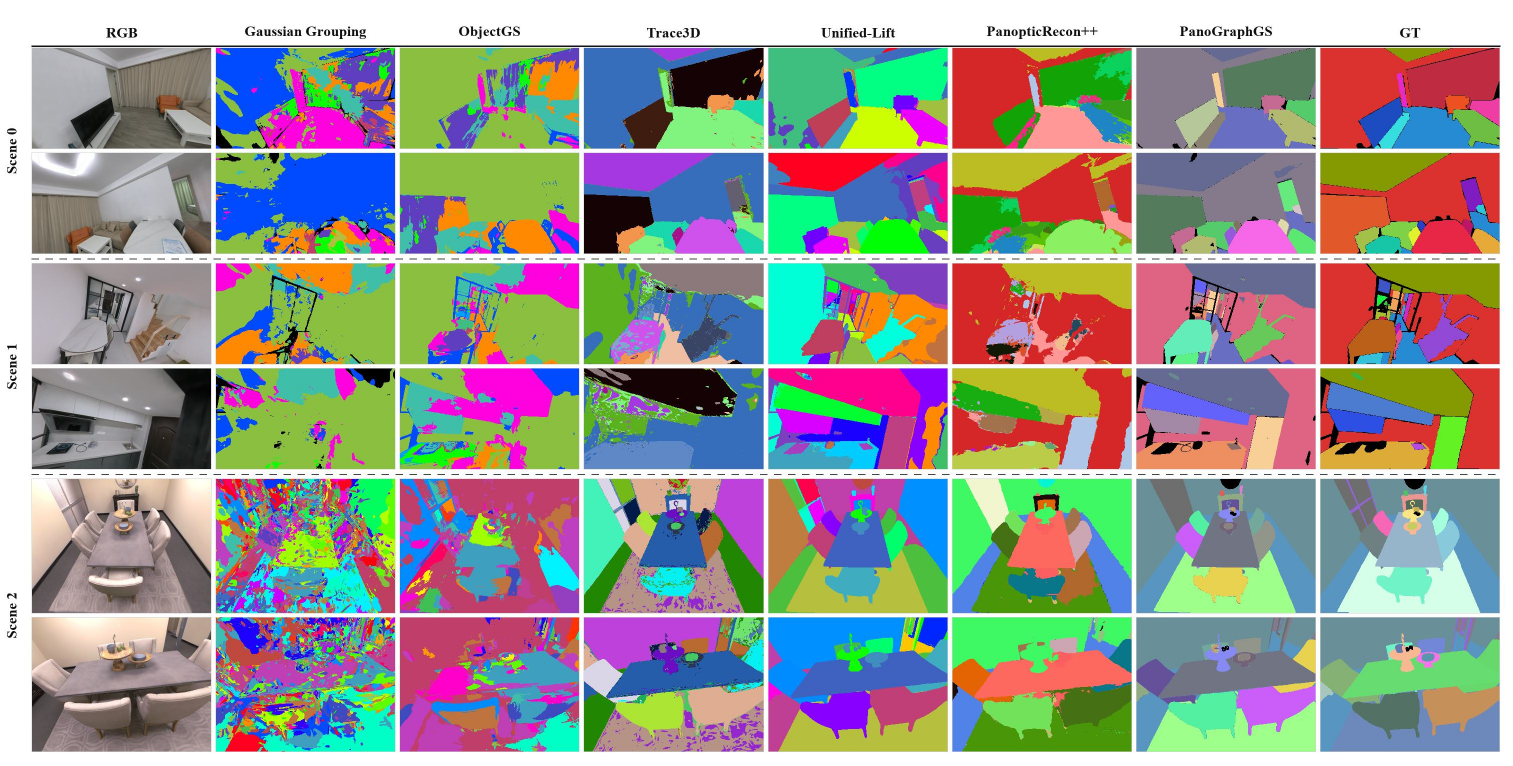}
    \caption{Panoptic segmentation visualization on NaviGS
    and Replica dataset.}
    \label{fig:qualitative_comparison}
\end{figure*}

\subsection{Experimental Setup}

\textbf{Datasets.}
We evaluate on two public datasets,
ScanNet++~\cite{yeshwanth2023scannetpp} and
Replica~\cite{straub2019replica}. In addition, we release \textbf{NaviGS},
a real-world dataset comprising RGB and semantic scenes. NaviGS is acquired
using two fisheye cameras and a handheld LiDAR device. LiDAR-assisted
acquisition provides accurate camera poses and a sparse point cloud for
Gaussian initialization. Both scene collections provide 3DGS models, meshes,
RGB, camera poses, and depth; semantic scenes additionally include
scene graphs.
To evaluate hierarchical scene graph construction, we select multi-room layouts from Replica and NaviGS that support
evaluation of the floor-room-stuff-object hierarchy.

\textbf{Evaluation Metrics.}
Scene graph performance is evaluated by floor-, room-, and object-level
accuracy. All compared methods use the same SAM3-generated segmentation masks
as supervision. Since OP3DSG, BBQ, and ConceptGraphs do not support
floor-room recognition, they are evaluated only at the object level. For
KeySG, HOV-SG, and PanoGraphGS, floor, room, and object recognition are
manually verified under a hierarchical criterion: a room is correct only when
its floor and room identities and semantic category are correctly identified,
while an object is correct only when its floor, room, and object identities
are all correct. Panoptic segmentation is evaluated using mIoU for region
overlap and mBIoU for object and structural boundaries. For instance-only
methods such as Unified-Lift, these metrics are computed after per-image
optimal instance matching without semantic categories. For panoptic methods,
including PanopticRecon++ and PanoGraphGS, a match is correct only when both
instance identity and semantic category are correctly predicted, yielding a
stricter criterion.

\textbf{Implementation Details.}
Training and evaluation are performed on an RTX 5880 GPU using Python and
PyTorch. Each scene is optimized for 30,000 iterations with Adam at a learning
rate of $2.5 \times 10^{-3}$. The number of semantic prototypes is determined
by the categories present in each scene, whereas the instance branch maintains
$K=256$ prototypes, sufficient for all evaluated scenes.

\subsection{Comparative Evaluation}

\textbf{3D scene graph evaluation.}
Table~\ref{tab:scenegraph_three_metrics} compares PanoGraphGS with
representative scene graph methods, while Fig~\ref{fig:scenegraph_visualization}
shows qualitative comparisons on multi-room scenes from Replica and NaviGS.
All compared methods rely on per-pixel RGB-D observations, whereas PanoGraphGS
constructs scene graphs without depth input or depth projection. PanoGraphGS
achieves strong floor, room, and object recognition, with object-node accuracy
exceeding existing methods by more than 50\% and reaching 98.36\% on NaviGS.
Compared with HOV-SG~\cite{werby2024hovsg}, which also supports the
floor-room-stuff-object hierarchy, PanoGraphGS achieves higher floor and room
recognition accuracy, reducing cross-floor and cross-room association errors
and improving object recognition. The 100\% floor accuracy of HOV-SG on
Replica does not demonstrate multi-floor reasoning, as the evaluated scenes
contain no multi-floor layouts. These results highlight the effectiveness of
our Semantic-Aware Floor and Room Reasoning module, which combines structural
semantics with geometric cues rather than relying primarily on point-cloud
density, distance thresholds, or watershed-based segmentation.
HOV-SG also generates numerous false-positive stuff nodes. For example, in
Scene~1, it identifies only one floor and incorrectly labels walls and ceilings as objects. This further demonstrates the effectiveness of our
Prototype-Based Semantic Learning in distinguishing semantic categories.
Compared with ConceptGraphs~\cite{gu2023conceptgraphs}, a strong object
recognition method, PanoGraphGS substantially reduces missed detections,
background false positives, and object fragmentation. As shown in Scene~0,
ConceptGraphs produces fragmented and noisy predictions and divides a single
sofa into three objects, whereas PanoGraphGS correctly identifies it as one
sofa, highlighted in yellow. This demonstrates the effectiveness of
Association-Based Instance Learning in associating observations of the same
physical object with a persistent instance identity.
Overall, PanoGraphGS achieves the best scene graph performance among the
evaluated methods without depth projection. These results demonstrate the
effectiveness of integrating panoptic Gaussian representation learning with
hierarchical 3D scene graph construction for accurate scene-level structural
reasoning without depth-based geometric projection.

\textbf{2D panoptic evaluation.} Table~\ref{tab:main_quantitative} presents the quantitative comparison, while Figure~\ref{fig:qualitative_comparison} provides qualitative results. Compared with Unified-Lift~\cite{zhu2025unifiedlift}, the strongest baseline, PanoGraphGS achieves the best mIoU and mBIoU across all three datasets. Unified-Lift produces instance segmentation but lacks semantic labels for panoptic understanding. For example, Unified-Lift assigns different instance IDs to three wall regions in Scene 2. In contrast, PanoGraphGS jointly predicts semantic categories and persistent instance identities, producing complete panoptic segmentation with more consistent identities across viewpoints. Compared with PanopticRecon++~\cite{yu2025panopticreconpp}, the state-of-the-art panoptic segmentation method, PanoGraphGS substantially improves both quantitative accuracy and visual quality, producing more complete regions, more stable object identities, and clearer object and structural boundaries.
These results further demonstrate the effectiveness of the residual projection MLP shared by rendered 2D pixel features and 3D Gaussian-level features; this design addresses the limitation of conventional semantic Gaussian methods whose learning modules typically operate on either 2D or 3D features.

\begin{table}[!t]
\centering
\caption{Ablation study on Replica.}
\label{tab:scenegraph_ablation}
\renewcommand{\arraystretch}{1}
\setlength{\tabcolsep}{2.5pt}
\small
\resizebox{\linewidth}{!}{%
\begin{tabular}{c|ccc|ccc}
\toprule
 & \shortstack{Confidence\\Weighted} &
\shortstack{Instance\\Association} &
\shortstack{Semantic\\Prototypes} &
\shortstack{Object\\Acc. (\%)} &
\shortstack{2D\\mIoU} &
\shortstack{2D\\mBIoU}\\
\midrule
EXP0 & $\times $ & $\times$ & $\times$  & 53.20 & 59.34 & 48.19 \\
EXP1 & $\checkmark$ & $\times$ & $\checkmark$  & 55.31 & 63.84 & 54.10\\
EXP2 & $\checkmark$ & $\checkmark$ & $\times$ & 80.42 & 70.23 &61.00\\
EXP3 & $\times$ & $\checkmark$ & $\checkmark$ & 86.33 & 68.97 &59.53\\
EXP4 & $\checkmark$ & $\checkmark$ & $\checkmark$ & \textbf{89.90} & \textbf{76.73}& \textbf{70.51}\\
\bottomrule
\end{tabular}}
\end{table}

\subsection{Ablation Study}

We evaluate the effectiveness of the individual components of PanoGraphGS on the public Replica dataset, which contains multi-room layouts. The comprehensive results are reported in Table~\ref{tab:scenegraph_ablation}. With the introduction of the semantic branch, EXP1 achieves a 2.11\% improvement in object accuracy over EXP0, demonstrating the effectiveness of jointly optimizing shared features across the 2D and 3D domains and enhancing semantic discrimination across views. EXP2 further validates the necessity of the instance branch, achieving the largest performance improvement of 27.22\% in object accuracy. Compared with EXP0, it substantially reduces false positives and missed detections caused by inconsistent annotations and supervision across frames, highlighting the importance of aligning view-local instance IDs with persistent identities to maintain consistent instance regions and boundaries across views. EXP4 further introduces the Confidence-weighted Mechanism on top of EXP3, improving object accuracy by 3.57\%. Its gains are even more pronounced in 2D panoptic segmentation, with improvements of 7.76\% and 10.98\% in mIoU and mBIoU, respectively. These results demonstrate that the confidence-weighted mechanism suppresses noisy supervision and stabilizes labels near region boundaries at the semantic level, while filtering unreliable identity targets and reducing conflicts around instance boundaries at the instance level.

\section{CONCLUSION}

In this study, we present PanoGraphGS, a unified framework integrating panoptic 3D Gaussian representation learning with hierarchical 3D scene graph construction. Prototype-based semantic learning jointly optimizes shared 2D and 3D features to mitigate noisy semantic supervision, while association-based instance learning aligns view-local masks with persistent object identities. The learned panoptic Gaussian field supports novel-view 2D panoptic segmentation and direct Gaussian-level decoding, while Semantic-Aware hierarchical 3D scene graph construction enables a semantically and geometrically grounded floor-room-stuff-object hierarchy. Extensive experiments show that PanoGraphGS achieves state-of-the-art performance among the evaluated methods in both 2D panoptic segmentation and 3D scene graph construction. We further release NaviGS with RGB and semantic scene data for panoramic scene understanding. PanoGraphGS nevertheless has several limitations. Open-plan spaces such as open kitchens lack explicit spatial boundaries, making room delineation challenging. Moreover, the current scene graph construction operates offline, and we will investigate online feed-forward scene graph construction for incremental scene understanding during continuous observations.


\FloatBarrier
\bibliographystyle{IEEEtran}
\bibliography{reference}

\end{document}